\documentclass[sigconf]{acmart}

\AtBeginDocument{%
  \providecommand\BibTeX{{%
    \normalfont B\kern-0.5em{\scshape i\kern-0.25em b}\kern-0.8em\TeX}}}

\copyrightyear{2023}
\acmYear{2023}
\setcopyright{acmlicensed}
\acmConference[ICISDM 2023]{2023 the 7th International Conference on Information System and Data Mining (ICISDM)}{May 10--12, 2023}{Atlanta, USA}
\acmBooktitle{2023 the 7th International Conference on Information System and Data Mining (ICISDM) (ICISDM 2023), May 10--12, 2023, Atlanta, USA}
\acmPrice{15.00}
\acmDOI{10.1145/3603765.3603771}
\acmISBN{979-8-4007-0063-7/23/05}

\usepackage{subcaption}
\usepackage{algorithm2e}

\SetKwComment{Comment}{/* }{ */}
\SetKwInput{KwInput}{Input}                
\SetKwInput{KwOutput}{Output}              
\RestyleAlgo{ruled}

\begin{document}

\title[SimpleGermKG]{Applying BioBERT to Extract Germline Gene-Disease Associations for Building a Knowledge Graph from the Biomedical Literature}

\author{Armando D. Diaz Gonzalez}

\email{addiazgonzalez@csustudent.net}
\orcid{0000-0002-4061-1880}
\affiliation{%
  \institution{Charleston Southern University}
  \streetaddress{9200 University Blvd}
  \city{North Charleston}
  \state{South Carolina}
  \country{USA}
  \postcode{29406}
}

\author{Songhui Yue}
\orcid{0009-0002-0945-3105}
\affiliation{%
  \institution{Charleston Southern University}
  \streetaddress{9200 University Blvd}
  \city{North Charleston}
  \state{South Carolina}
  \country{USA}
  \postcode{29406}
  }
\email{syue@csuniv.edu}

\author{Sean T. Hayes}
\orcid{0000-0003-3631-7782}
\affiliation{%
  \institution{Charleston Southern University}
  \streetaddress{9200 University Blvd}
  \city{North Charleston}
  \state{South Carolina}
  \country{USA}
  \postcode{29406}
}
\email{shayes@csuniv.edu}

\author{Kevin S. Hughes}
\orcid{0000-0003-4084-6484}
\affiliation{%
  \institution{Medical University of South Carolina}
  \streetaddress{171 Ashley Ave}
  \city{Charleston}
  \state{South Carolina}
  \country{USA}
  \postcode{29425}
 }
\email{hughkevi@musc.edu}

\renewcommand{\shortauthors}{Armando D. Diaz Gonzalez, et al.}

\begin{abstract}
Published biomedical information has and continues to rapidly increase. The recent advancements in Natural Language Processing (NLP), have generated considerable interest in automating the extraction, normalization, and representation of biomedical knowledge about entities such as genes and diseases. Our study analyzes germline abstracts in the construction of knowledge graphs of the immense work that has been done in this area for genes and diseases. This paper presents SimpleGermKG, an automatic knowledge graph construction approach that connects germline genes and diseases. For the extraction of genes and diseases, we employ BioBERT, a pre-trained BERT model on biomedical corpora. We propose an ontology-based and rule-based algorithm to standardize and disambiguate medical terms. For semantic relationships between articles, genes, and diseases, we implemented a part-whole relation approach to connect each entity with its data source and visualize them in a graph-based knowledge representation. Lastly, we discuss the knowledge graph applications, limitations, and challenges to inspire the future research of germline corpora. Our knowledge graph contains 297 genes, 130 diseases, and 46,747 triples. Graph-based visualizations are used to show the results.
\end{abstract}





\maketitle

\section{Introduction}
Certain genes that a person is born with protect us from developing cancer. Cancer susceptibility have mutations (i.e., have a DNA change that prevents their normal function), creating a higher risk of developing cancer. Looking for which mutated genes increase the risk of which specific cancers is of great interest and is known as the {\itshape gene-disease association}. Extracting germline genes and diseases from biomedical corpora for representing knowledge encoded in a {\itshape Knowledge Graph} (KG) requires complex, expensive, and time-consuming methods. Biomedical publications are increasing rapidly. For example, using the same search criteria, a PubMed search for BRCA1 and BRCA2 in 2010, fetched 478 papers compared to 830 papers in 2021, a 57\% increase in annual new papers in 11 years. Because the total number of publications on these genes now exceeds 12,300 and there are an estimated 22,287 genes in the human genome \cite{1}, the magnitude of this task becomes overwhelming. As a result, manual extraction is essentially impossible. Many computational approaches have been proposed to extract gene-disease association information from the biomedical literature accurately and efficiently. For instance, in the fields of pharmacy \cite{2}, medicine \cite{3}, and biology \cite{4}, machine learning and deep learning models have enabled biomedical text-mining tasks such as summarizing, extracting, and analyzing large corpora with varying degrees of success \cite{5}.

Natural Language Processing (NLP), a field of artificial intelligence, is used to perform tasks such as {\itshape Named Entity Recognition} (NER), {\itshape Named Entity Normalization} (NEN), and {\itshape Relation Extraction} (RE) \cite{6,7,8,9,10}. NLP systems can analyze immense amounts of text-based data and determine the correct meaning of a word in a specific context to extract key facts and relationships. To address the problem of gene-disease associations in an article, NER can be used for extracting genes and diseases (as entities) from the biomedical corpora \cite{11}. The most recent approach is driven by transformer-based models that were recently developed by Google \cite{12} and can be used for carrying out various NLP tasks \cite{13}. This approach can be pre-trained on biomedical literature and is known to outperform pre-trained models, such as ELMo and BERT \cite{14}.

Unlike relational databases, graph databases provide unique abilities to manage n-th degree relationships among complex types of biomedical data \cite{15}. Knowledge Graphs (KGs) have proven to be effective in representing large-scale heterogeneous data and visualizing the nature of underlying relationships. KGs provide a model of relevant facts and contextualized answers to specific questions, so that they can then be used to extract and discover deeper and more subtle patterns \cite{16}. For example, KGs are suitable for representing hierarchical data, such as genes, diseases, and relationships that are interconnected.

Furthermore, many studies focus on a particular segment in the three-stage life cycle of the knowledge graph construction process that includes NER, NEN, and RE or Semantic Relation. In this paper, we present SimpleGermKG, a gene-disease knowledge graph based on germline corpora. The germline genes and diseases are extracted from abstracts using BioBERT \cite{14}. To our knowledge, no study has been conducted to analyze germline abstracts in the construction of knowledge graphs. Therefore, we examine the knowledge graph life cycle based on a hybrid procedure between deep learning, ontology-based, and rule-based approaches beginning with the data pre-processing, knowledge graph construction part, and ending with a discussion of graph applications and visualizations for further analysis.

Our contributions are summarized as follows:
\begin{itemize}
\item {We automated the construction of SimpleGermKG, which visually organizes genes and diseases from germline abstracts. SimpleGermKG will expedite searches for gene-disease associations with references.}
\item {We developed SimpleGermKG using BioBERT to extract genes and diseases from biomedical texts. Then, an ontology-rule-based NEN algorithm was designed to match genes and diseases with master terms. Lastly, a part-whole relation approach to connect these gene-disease pairs with their references.}
\item {In Section 5, we proposed three relationship approaches for classifying relationships between germline genes and diseases. Two of them are based on a co-occurrence method, which indicates that there is a possible relationship between two entities when these appear in the text. The last approach could be used to find more granular relationships using a pre-trained language model such as BioBERT.}
\item {The source code of our workflow is freely available at \url{https://github.com/arm-diaz/Bio-Germline-Diseases-BERT-NER.}
}
\end{itemize}

The structure of this paper is as follows: Section 2 presents an overview of relevant approaches to the biomedical knowledge graph life cycle in previous studies. Section 3 explains a general description of the proposed methodological approach. Section 4 describes details of the developed workflow, and the case study results. Section 5 discusses future work. Finally, the conclusions are highlighted in Section 6.

\section{Related Work}

Biomedical knowledge graphs may be constructed using various techniques which begin with large datasets that are extracted from pre-existing databases or texts. These pre-existing databases were created by domain experts using manually curated KGs and automatically extracted KGs (e.g., using machine learning methods). Manual curation is a time-consuming process, due to the required effort by the domain expert to review papers, annotate phrases and sentences, and define rules and constraints that help users make inferences. On the other hand, machine learning approaches in natural language processing tasks can be used to automate the process of building a knowledge graph. NLP can quickly detect sentences of interest and unveil complex relationships among the data. These methods require annotating only a subset of the data.

\subsection{Ontology-based Knowledge Graph Construction}

In medicine, biomedical knowledge can be divided into many subdomains, such as genes, chemical compounds, diseases, organs, symptoms, and syndromes. The purpose of biomedical ontology goes beyond collecting names of entities, a dictionary of terms, and controlled vocabulary for a variety of entities. It defines biological classes of entities and the relations among them for building a knowledge base \cite{17}. A well-defined ontology is essential for the creation of a biomedical knowledge graph because the ontology enables complex reasoning about biomedical knowledge. Some of these KGs, e.g. GARD \cite{15} and GenomicKB \cite{18} have made significant contributions to the integration and utilization of existing biomedical knowledge regarding rare disease information sources, human genome, epigenome, transcriptome, and 4D nucleome to provide patients with the latest health information to answer human genomics-related questions.

BioPortal \cite{19}, an open repository of biomedical ontologies, has more than 1,000 ontologies and 15 million classes of entities. These ontologies have been designed and developed by the community of research teams to summarize and organize information. Maintaining an ontology through its life cycle is infeasible for a human expert since it is expensive and time-consuming. In addition, the difficulty is compounded by the fact that high-quality and scalable ontologies require reusing parts of other ontologies and applying automated quality control testing that guarantees best practices for software development \cite{20}.

\subsection{Automatic Knowledge Graph Construction}

Managing the increased rate of publications via manual curation is infeasible, requiring approaches that can automate part or all of the process. Natural Language Processing is commonly used to extract entities and their relations from biomedical text. Therefore, NLP can facilitate and automate knowledge graph construction. NER and NEN approaches have been developed to find relevant entities and connect these entities to meet the agreed data model \cite{21}. Biomedical named entity recognition and named entity resolution techniques have been studied since the late 1990s \cite{22}, and different approaches have been proposed and developed to solve NER systems. These approaches can be classified into (1) rule-based, which relies on linguistic experts designing accurate rules, (2) machine learning-based, such as Hidden Markov Models (HMM) and Conditional Random Fields (CRF), (3) deep learning-based such as RNN, LSTM, CNN, and pre-trained language models, and (4) hybrid approaches \cite{23,24,25,26}.

Due to significant advances in deep learning, pre-training allows the model to incorporate domain-specific knowledge, which can further improve the ability of the pre-trained base model to achieve high performance on various tasks \cite{26}.  The model can be fine-tuned on task-specific datasets to perform tasks, such as named entity recognition and relation extraction \cite{14,27,28}, which are two critical tasks to construct domain-specific knowledge graphs. In contrast with a manual curation approach, pre-trained models can reduce computing costs, and save time and resources. For example, BERT-based models can be used to generate scalable knowledge graphs from new corpora that include undiscovered knowledge \cite{29}.

A recent study, HerbKG \cite{30}, a knowledge graph that bridges herbal and molecular medicine, uses text mining techniques, such as PubTator Central (PTC) NER model and a custom BERT-based RE model to produce a list of identified relation triplets, which are used for the HerbKG construction. The constructed HerbKG supports multiple downstream applications, such as descriptive analysis, evidence-based graph query, similarity analysis, and drug repurposing. Other studies \cite{21,31} have attempted to apply machine learning algorithms and BERT-based models to extract relationships, such as Drug-Gene, Drug-Disease, and Gene-Disease, and have a clearer understanding of diseases, symptoms, and gene mutations.

\section{Method}

\begin{figure}[h]
  \centering
  \includegraphics[width=\linewidth]{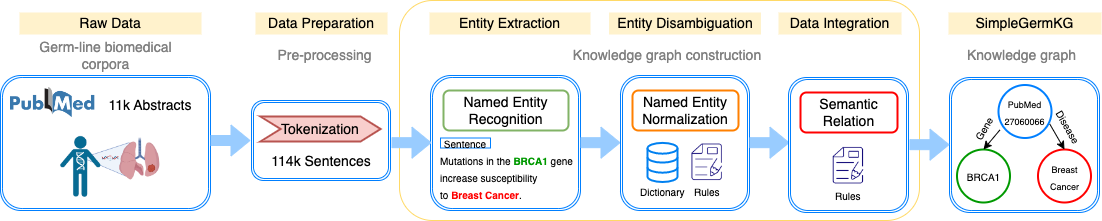}
  \caption{SimpleGermKG Architecture. This figure illustrates the overall workflow of SimpleGermKG. BioBERT is pre-trained on PubMed abstracts to extract germline genes and diseases. Then, ambiguated entities are eliminated and a semantic relation approach is utilized to build a knowledge graph (KG) and improve the interpretability of our results.}
  \Description{SimpleGermKG Architecture.}
  \label{fig:simplegermkg}
\end{figure}

We propose a four-stage pipeline, which constructs SimpleGermKG. First, we proceed with a detailed description of the dictionaries used for the NEN task. Then, we describe the workflow that consists of tokenizing and preparing the dataset for machine learning, extracting genes and diseases from germline corpora using BioBERT NER, standardizing entities through a named entity normalization process, and linking normalized entities using the semantic relation that associates entities with their PubMed ID as illustrated in Figure \ref{fig:simplegermkg}.

\subsection{Data Sources}

Due to the complexity of properly defining and categorizing a large number of biomedical terms, we rely on two home-grown dictionaries (MUSC), one includes a list of diseases, and the other a list of genes. These dictionaries are relevant for mapping genes and diseases to a master term. The dictionary of diseases contains around 125 disease names and 452 synonyms, and the dictionary of genes contains around 336 gene names and 1,310 synonyms.

\subsection{Pre-processing}

Tokenization is the process of breaking down unstructured data and natural language text into smaller units of information \cite{32}. For instance, sentences, punctuation marks, words, and numbers can be considered tokens. Large input sizes for machine learning models are not recommended, especially BERT-based models \cite{26}, which have an input size restriction of 512 characters. Considering an abstract of a research paper is usually a paragraph of 300 words or less, an abstract may have over 512 characters with spaces included in the character count. To solve this problem, we used the PunktSentenceTokenizer \cite{33} method from the NLTK python library, which is trained on the Penn Treebank corpus and uses regular expressions to parse sentences and detect the sentence boundaries.

\subsection{Named Entity Recognition}

We used a fine-tuned BioBERT model pre-trained on NCBI-disease corpus \cite{34} to extract diseases from germline abstracts. The Natural Center for Biotechnology Information Disease (NCBI) \cite{35} disease corpus is a collection of 793 PubMed articles with 6,892 manually annotated disease mentions. For extracting genes from germline abstracts, we used a fine-tuned BioBERT model pre-trained on the BC2GM corpus \cite{36}. The BioCreative II Gene Mention (BC2GM) corpus \cite{37} consists of sentences from PubMed abstracts with manually labeled gene and alternative gene entities.

\subsection{Named Entity Normalization}

After extracting single and multi-word phrases from texts, Named Entity Normalization \cite{38} is performed, which allocates suitable tags to the recognized entities. In biomedical articles, named entity normalization is a challenging task because biological terms, such as genes and diseases have multiple synonyms, and term variations, and are often referred to using abbreviations \cite{39}. To resolve these ambiguities, machine-learning approaches \cite{40} have been investigated. However, many normalization tools rely on domain-specific ontologies, dictionaries, or rules. Domain-specific dictionaries can differentiate between synonyms, abbreviations, and punctuation marks.

We used a dictionary-lookup approach using our two manually curated dictionaries and an approximate string-matching algorithm. The algorithm converts identified entities from text to lexical variations, such as lowercasing and removing whitespace and punctuations, and then maps them to specific master terms. To reduce the complexity of our Algorithm \ref{alg:norm}, we assume the BioBERT NER task classifies a single entity following the format BIO encoding scheme to represent the tokens and each entity that may be composed of several words must be unique. For example, the input sequence ``BRCA1 and BRCA2'' should be classified with the labels B-GENE, I-GENE, and B-GENE. This assumption allows us to map only one master term from the dictionary for each entity. However, an entity may be mapped to more than one master term because the BioBERT NER task may classify the previous example with the labels B-GENE, I-GENE, and I-GENE.

\begin{algorithm}[h]
\caption{Named Entity Normalization}\label{alg:norm}
  \KwInput{Ontology Set, $\textit{O}=\{{\textit{O}_{1},\textit{O}_{2},...,\textit{O}_{n}}$\}
            \linebreak Named Entity Set, $\textit{E}=\{{\textit{E}_{1},\textit{E}_{2},...,\textit{E}_{n}}$\}}
  \KwOutput{Disambiguated Entity Set, $\textit{D}=\{{\textit{D}_{1},\textit{D}_{2},...,\textit{D}_{n}}$\}}
$O (i, j) \gets $Ontology Set$  $ \,\,$/*$[r]{Ontology of genes and diseases}\,\,$/*$\\
$E (i, j) \gets $Named Entity Set$   $ \,\,$/*$[r]{Named Entity}\,\,$/*$\\
$D (i, j) \gets $\O$    $ \,\,$/*$[r]{Disambiguated Entity}\,\,$/*$\\
\For{all elements ($index$, $entity$) in $E (i, j)$}{
  \uIf{$entity$ in O}{
    $D(index, entity) \gets O(entity)$;
  }\uElseIf{StringMatch($entity$,  $O$)}{
      $D(index, entity) \gets StringMatch($entity$,  $O$)$;
  }\Else{
    $D(index, entity) \gets \emptyset$;
  }
}
\end{algorithm}

\subsection{Semantic Relation}

Given a pair of entities, such as a gene and disease, a semantic relation consists of identifying the relation type between them. An important semantic relation for many applications is the part-whole relation \cite{41}. Let us notate the part–whole relation as PART (<Tail Entity>, <Head Entity>), where <Tail Entity> is part of <Head Entity>. For example, the phrase ``genes are found on tiny structures called chromosomes'' contains the part-whole relation PART (genes, chromosomes). More recent studies, such as the SemEval 2018 Task 7 \cite{42}, proposed a task on semantic relation extraction and classification in scientific paper abstracts that are practical for working on extracting specialized knowledge from domain corpora, such as biomedical information extraction.

Successful entity-relation linking requires detecting both the entity mentions in the abstracts, along with their respective entity types from the gene-disease dictionaries, and determining the type of relationship that exists between them. Based on psycholinguistic experiments and how the entities contribute to the structure of the part-whole relationship, we determined that the part-whole relationship from SemEval 2018 Task 7 can help us better identify and connect our entities to build the knowledge graph. SemEval 2018 Task 7 provides three comprehensive sets of classification rules, (1) composed of, (2) data source, and (3) phenomenon \cite{42}. 

Our main dataset contains germline abstracts and their PubMed ID. An abstract can include more than one gene and disease mentioned per sentence. Because germline mutations are passed on from parents to offspring, it is complicated to establish causal relationships in the germline association between genes and diseases, and thus they are not well-defined in the literature \cite{43}. Therefore, we use a data source relationship in the form of PUBMED\_ID-GENES\_IN-GENE and PUBMED\_ID-DISEASES\_IN-DISEASE. Our approach matches all given genes and diseases to their given PubMed ID.

\section{Results}

\subsection{Knowledge Graph Construction}

Our experiments are conducted on germline corpora, which contain 11,261 abstracts from PubMed, and 114,311 sentences after tokenization. The BioBERT-NER approach identified 19,751 gene entities and 19,135 disease entities. We also detected that most of the gene and disease entities are synonyms and therefore, are part of the same entity. To eliminate the data ambiguities, we applied an {\itshape ad hoc} mapping and filtering procedure (Algorithm 1). We also eliminated entities that represent objects that did not match our ontology. Then, we formally defined a semantic relation type to be a pair consisting of a domain class of type PART-DATASOURCE (Gene, PubMed ID) and PART-DATASOURCE (Disease, PubMed ID). We defined the semantic relation as ``GENES\_IN'', and ``DISEASES\_IN'' to capture the connection between a PubMed ID and genes and/or diseases. Once we identified the disambiguated entity types and relationships, we linked them together and built the knowledge graph. The knowledge graph was built with the Neo4j graph platform and contains 46,747 triples with 9,414 entities, including 8,987 PubMed IDs, 297 genes, and 130 diseases.

BioPortal \cite{19}, an open repository of biomedical ontologies, has more than 1,000 ontologies and 15 million classes of entities. These ontologies have been designed and developed by the community of research teams to summarize and organize information. Maintaining an ontology through its life cycle is infeasible for a human expert since it is expensive and time-consuming. In addition, the difficulty is compounded by the fact that high-quality and scalable ontologies require reusing parts of other ontologies and applying automated quality control testing that guarantees best practices for software development \cite{20}.

\subsection{Knowledge Graph-based Visualization}

\begin{figure}
     \centering
     \begin{subfigure}[b]{0.21\textwidth}
         \centering
         \includegraphics[width=\linewidth]{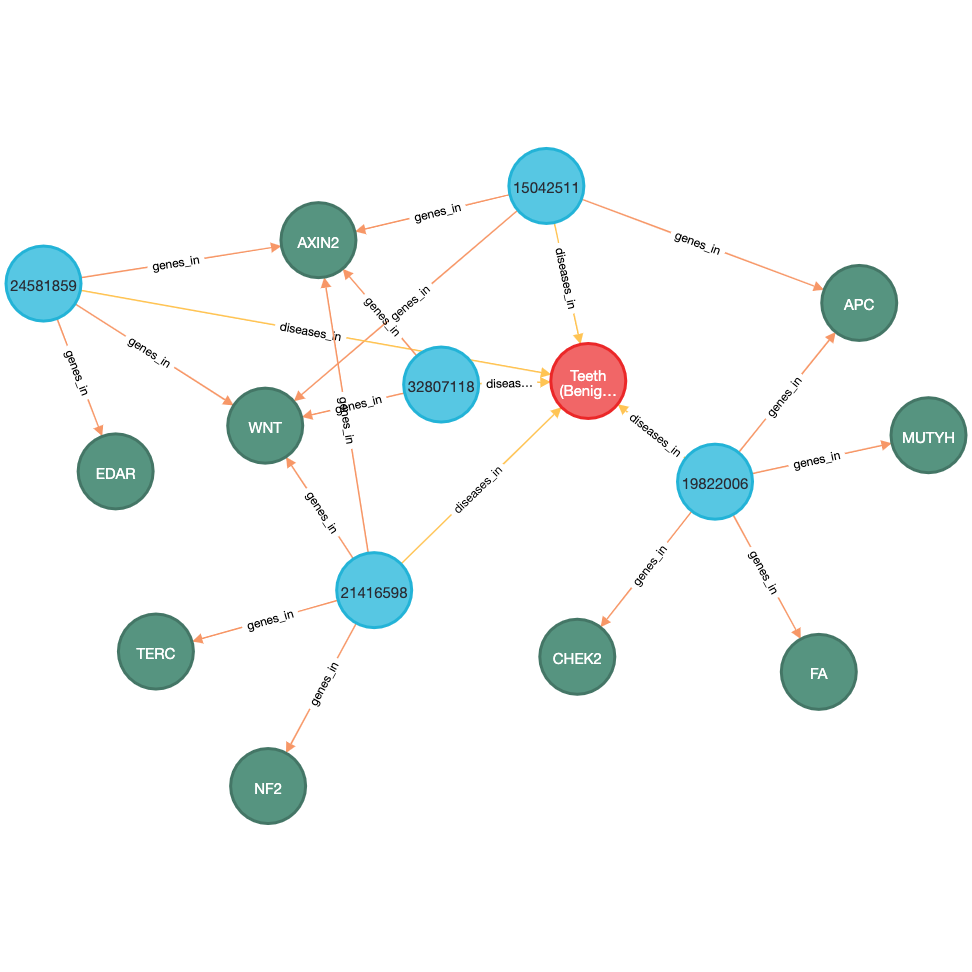}
         \caption{PubMed IDs \& ``Teeth (Benign)'' \& Genes}
         \Description{Graph a) shows articles that mentioned the disease ``Teeth (Benign)'' and all gene entities mentioned in the abstract}
         \label{fig:grapha}
     \end{subfigure}
     \hfill
     \begin{subfigure}[b]{0.21\textwidth}
         \centering
         \includegraphics[width=\linewidth]{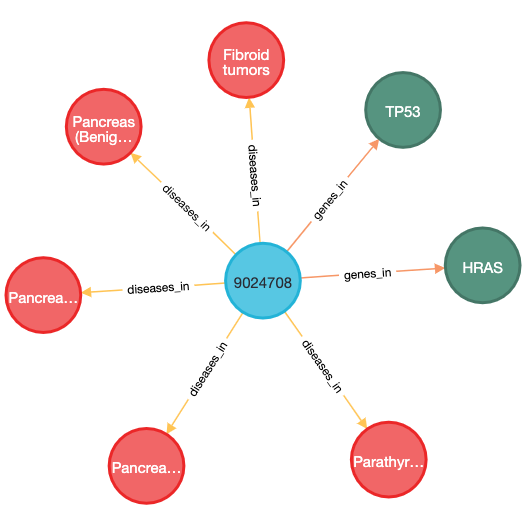}
         \caption{``9024708'' \& Diseases \& Genes}
         \label{fig:graphb}
     \end{subfigure}
        \caption{Graph Representation of Gene-PubMed-disease Associations. Graphs are generated by querying the knowledge graph stored in Neo4j}
        \Description{Graph b) shows genes and diseases mentioned in the abstract of the article ``9024708''}
      \label{fig:graphs}
\end{figure}

So far, SimpleGermKG which covers germline abstracts from PubMed has been constructed with an integrated ontology of genes and diseases. We store the SimpleGermKG in the Neo4j graph database, which allows researchers and clinicians to find relevant information and facilitates the navigation of biomedical data. To prove the visual management and ease-to-query of Neo4j, we show the search results of two queries through the Neo4j Cypher graph query language in Figure \ref{fig:graphs}. The blue node represents ``PubMed ID'' (abstract ID), the green node is ``gene'' (disambiguated gene names), and the red node is ``disease'' (disambiguated diseases) mentioned in the text. The ``gene'' and ``disease'' nodes can be also identified with the relationship (edge) ``GENES\_IN'' and ``DISEASES\_IN'' respectively. Figure \ref{fig:grapha} shows articles that mentioned the disease ``Teeth (Benign)'' and all gene entities mentioned in the abstracts. Figure \ref{fig:graphb} shows genes and diseases mentioned in the abstract of the article ``9024708''.

\section{Discussion and Future Work}

The construction of knowledge graphs on germline corpora presents new opportunities as few studies focus on this domain. SimpleGermKG has the potential to integrate information from electronic health records, genomic data, and other existing biomedical ontologies. Our knowledge graph has improved the search capabilities of medical practitioners by helping them to retrieve relevant research papers for a particular disease or condition and genetic variations. As more information is added to SimpleGermKG, we expect to broaden its applications. Some of the future activities on utilizing and improving our SimpleGermKG will involve:

\begin{itemize}
\item {Exploring possible applications and opportunities that could improve the lifestyle of individuals who present germline mutations. Germline mutations may affect people differently depending on genetic factors such as family background. These mutations may present a certain level of resistance to the effects of drugs. Therefore, it is important to explore opportunities in patients and identify possible risks, therapies, and clinical implications.}
\item {Experimenting and exploring state-of-the-art approaches for the NER task. We aim to improve the precision of the gene-disease extraction by exploring pre-trained language models that have been fine-tuned on well-known gene and disease datasets in the literature.}
\item {Exploring a method of expanding our dictionaries for the NEN task. Larger gene-disease ontologies can be explored to enrich the vocabulary and improve the accuracy of the named entity normalization process. We can rely on other ontologies by combining concepts to generate a more complete vocabulary that includes more variations of the same terms from biomedical texts.}
\item {Developing a technique for obtaining relationships from germline corpora. Due to the nature of germline mutations, conventional relation extraction techniques do not apply in the semantic relation of a germline corpus. Therefore, training a model on germline corpora should consider the gene carrier probability to select risk families for extracting relationships between cancer susceptibility genes. We propose three methods to identify the presence of associations between genes and diseases
}: \begin{itemize}{
\item {Article level - Mentions of genes and diseases in the same article have a direct relationship with the PubMed ID. This approach cannot directly link a gene and disease. But we know that those entities have a contextual relationship within the text.}
}
\item {Sentence level - In contrast to the article-based approach, we can assume a relationship between a gene and disease exists when those are mentioned in the same PubMed ID and sentence ID.}
\item {BERT-based approach - A relation classification approach can extract sentences that contain the entity pair from the NER task which holds a semantic relation and then predicts whether a certain relation exists between genes and diseases.}
\end{itemize}
\end{itemize}

\section{Conclusion}

Digital biomedical information has been growing exponentially. To represent biomedical information effectively, we developed an automated knowledge graph construction framework, SimpleGermKG, to synthesize and store detailed information about genes and diseases associated with a PubMed ID. We employed BioBERT, a natural language processing model, to retrieve key information. A NEN algorithm was proposed to eliminate disambiguation. Our SimpleGermKG contains 297 genes, 130 diseases, and 46,747 triples. The knowledge graph can store and represent medical knowledge from large biomedical corpora in such a way that researchers, students, and physicians can search, manage, share, and visualize.

\begin{acks}
We appreciate the support of the senior researcher on this project, Dr. Kevin Hughes along with the Medical University of South Carolina, who provided the project data and effective guidance from project inception to completion. We also acknowledge assistance from our faculty mentors, Dr. Songhui Yue and Dr. Sean Hayes, in the development of this project and critical manuscript feedback.
\end{acks}

\bibliographystyle{ACM-Reference-Format}
\bibliography{biobliography}


\begin{thebibliography}{43}


\ifx \showCODEN    \undefined \def \showCODEN     #1{\unskip}     \fi
\ifx \showDOI      \undefined \def \showDOI       #1{#1}\fi
\ifx \showISBNx    \undefined \def \showISBNx     #1{\unskip}     \fi
\ifx \showISBNxiii \undefined \def \showISBNxiii  #1{\unskip}     \fi
\ifx \showISSN     \undefined \def \showISSN      #1{\unskip}     \fi
\ifx \showLCCN     \undefined \def \showLCCN      #1{\unskip}     \fi
\ifx \shownote     \undefined \def \shownote      #1{#1}          \fi
\ifx \showarticletitle \undefined \def \showarticletitle #1{#1}   \fi
\ifx \showURL      \undefined \def \showURL       {\relax}        \fi
\providecommand\bibfield[2]{#2}
\providecommand\bibinfo[2]{#2}
\providecommand\natexlab[1]{#1}
\providecommand\showeprint[2][]{arXiv:#2}

\bibitem[Abreu~Vicente(2022a)]%
        {36}
\bibfield{author}{\bibinfo{person}{J Abreu~Vicente}.}
  \bibinfo{year}{2022}\natexlab{a}.
\newblock \bibinfo{title}{{drAbreu}/{bioBERT}-{NER}-{BC2GM}\_corpus}.
\newblock
\newblock
\urldef\tempurl%
\url{https://huggingface.co/drAbreu/bioBERT-NER-BC2GM_corpus}
\showURL{%
\tempurl}


\bibitem[Abreu~Vicente(2022b)]%
        {34}
\bibfield{author}{\bibinfo{person}{J Abreu~Vicente}.}
  \bibinfo{year}{2022}\natexlab{b}.
\newblock \bibinfo{title}{{drAbreu}/{bioBERT}-{NER}-{NCBI}\_disease}.
\newblock
\newblock
\urldef\tempurl%
\url{https://huggingface.co/drAbreu/bioBERT-NER-NCBI_disease}
\showURL{%
\tempurl}


\bibitem[Al-Garadi et~al\mbox{.}(2022)]%
        {5}
\bibfield{author}{\bibinfo{person}{Mohammed~Ali Al-Garadi},
  \bibinfo{person}{Yuan-Chi Yang}, {and} \bibinfo{person}{Abeed Sarker}.}
  \bibinfo{year}{2022}\natexlab{}.
\newblock \showarticletitle{The {Role} of {Natural} {Language} {Processing}
  during the {COVID}-19 {Pandemic}: {Health} {Applications}, {Opportunities},
  and {Challenges}}.
\newblock \bibinfo{journal}{\emph{Healthcare}} \bibinfo{volume}{10},
  \bibinfo{number}{11} (\bibinfo{year}{2022}).
\newblock
\showISSN{2227-9032}
\urldef\tempurl%
\url{https://doi.org/10.3390/healthcare10112270}
\showDOI{\tempurl}


\bibitem[Al-Moslmi et~al\mbox{.}(2020)]%
        {16}
\bibfield{author}{\bibinfo{person}{Tareq Al-Moslmi}, \bibinfo{person}{Marc
  Gallofré~Ocaña}, \bibinfo{person}{Andreas L.~Opdahl}, {and}
  \bibinfo{person}{Csaba Veres}.} \bibinfo{year}{2020}\natexlab{}.
\newblock \showarticletitle{Named {Entity} {Extraction} for {Knowledge}
  {Graphs}: {A} {Literature} {Overview}}.
\newblock \bibinfo{journal}{\emph{IEEE Access}}  \bibinfo{volume}{8}
  (\bibinfo{year}{2020}), \bibinfo{pages}{32862--32881}.
\newblock
\urldef\tempurl%
\url{https://doi.org/10.1109/ACCESS.2020.2973928}
\showDOI{\tempurl}


\bibitem[Alshaikhdeeb and Ahmad(2016)]%
        {6}
\bibfield{author}{\bibinfo{person}{Basel Alshaikhdeeb} {and}
  \bibinfo{person}{Kamsuriah Ahmad}.} \bibinfo{year}{2016}\natexlab{}.
\newblock \showarticletitle{Biomedical {Named} {Entity} {Recognition}: {A}
  {Review}}.
\newblock \bibinfo{journal}{\emph{International Journal on Advanced Science,
  Engineering and Information Technology}} \bibinfo{volume}{6},
  \bibinfo{number}{6} (\bibinfo{year}{2016}), \bibinfo{pages}{889--895}.
\newblock
\showISSN{2088-5334}
\urldef\tempurl%
\url{https://doi.org/10.18517/ijaseit.6.6.1367}
\showDOI{\tempurl}
\newblock
\shownote{Publisher: INSIGHT - Indonesian Society for Knowledge and Human
  Development}.


\bibitem[Beltagy et~al\mbox{.}(2019)]%
        {27}
\bibfield{author}{\bibinfo{person}{Iz Beltagy}, \bibinfo{person}{Kyle Lo},
  {and} \bibinfo{person}{Arman Cohan}.} \bibinfo{year}{2019}\natexlab{}.
\newblock \showarticletitle{{SciBERT}: {A} {Pretrained} {Language} {Model} for
  {Scientific} {Text}}.
\newblock  (\bibinfo{year}{2019}).
\newblock
\urldef\tempurl%
\url{https://doi.org/10.48550/ARXIV.1903.10676}
\showDOI{\tempurl}
\newblock
\shownote{Publisher: arXiv}.


\bibitem[Bhatnagar et~al\mbox{.}(2022)]%
        {13}
\bibfield{author}{\bibinfo{person}{Roopal Bhatnagar}, \bibinfo{person}{Sakshi
  Sardar}, \bibinfo{person}{Maedeh Beheshti}, {and} \bibinfo{person}{Jagdeep~T
  Podichetty}.} \bibinfo{year}{2022}\natexlab{}.
\newblock \showarticletitle{How can natural language processing help model
  informed drug development?: a review}.
\newblock \bibinfo{journal}{\emph{JAMIA Open}} \bibinfo{volume}{5},
  \bibinfo{number}{2} (\bibinfo{year}{2022}).
\newblock
\showISSN{2574-2531}
\urldef\tempurl%
\url{https://doi.org/10.1093/jamiaopen/ooac043}
\showDOI{\tempurl}


\bibitem[Bodenreider et~al\mbox{.}(2005)]%
        {17}
\bibfield{author}{\bibinfo{person}{Olivier Bodenreider}, \bibinfo{person}{Joyce
  Mitchell}, {and} \bibinfo{person}{A McCray}.}
  \bibinfo{year}{2005}\natexlab{}.
\newblock \showarticletitle{Biomedical ontologies}.
\newblock \bibinfo{journal}{\emph{Pacific Symposium on Biocomputing. Pacific
  Symposium on Biocomputing}}  \bibinfo{volume}{78} (\bibinfo{year}{2005}),
  \bibinfo{pages}{76--78}.
\newblock
\urldef\tempurl%
\url{https://doi.org/10.1142/9789812704856_0016}
\showDOI{\tempurl}


\bibitem[Bonifaci et~al\mbox{.}(2010)]%
        {43}
\bibfield{author}{\bibinfo{person}{Núria Bonifaci}, \bibinfo{person}{Bohdan
  Górski}, \bibinfo{person}{Bartlomiej Masojć}, \bibinfo{person}{Dominika
  Wokołorczyk}, \bibinfo{person}{Anna Jakubowska}, \bibinfo{person}{Tadeusz
  Dębniak}, \bibinfo{person}{Antoni Berenguer}, \bibinfo{person}{Jordi
  Serra~Musach}, \bibinfo{person}{Joan Brunet}, \bibinfo{person}{Joaquín
  Dopazo}, \bibinfo{person}{Steven~A Narod}, \bibinfo{person}{Jan Lubiński},
  \bibinfo{person}{Conxi Lázaro}, \bibinfo{person}{Cezary Cybulski}, {and}
  \bibinfo{person}{Miguel~Angel Pujana}.} \bibinfo{year}{2010}\natexlab{}.
\newblock \showarticletitle{Exploring the {Link} between {Germline} and
  {Somatic} {Genetic} {Alterations} in {Breast} {Carcinogenesis}}.
\newblock \bibinfo{journal}{\emph{PLOS ONE}} \bibinfo{volume}{5},
  \bibinfo{number}{11} (\bibinfo{year}{2010}), \bibinfo{pages}{1--8}.
\newblock
\urldef\tempurl%
\url{https://doi.org/10.1371/journal.pone.0014078}
\showDOI{\tempurl}
\newblock
\shownote{Publisher: Public Library of Science}.


\bibitem[Cariello et~al\mbox{.}(2021)]%
        {7}
\bibfield{author}{\bibinfo{person}{Maria~Carmela Cariello},
  \bibinfo{person}{Alessandro Lenci}, {and} \bibinfo{person}{Ruslan Mitkov}.}
  \bibinfo{year}{2021}\natexlab{}.
\newblock \showarticletitle{A {Comparison} between {Named} {Entity}
  {Recognition} {Models} in the {Biomedical} {Domain}}.
  \bibinfo{publisher}{INCOMA Ltd.}, \bibinfo{address}{Held Online},
  \bibinfo{pages}{76--84}.
\newblock
\urldef\tempurl%
\url{https://aclanthology.org/2021.triton-1.9}
\showURL{%
\tempurl}


\bibitem[Chithrananda et~al\mbox{.}(2020)]%
        {28}
\bibfield{author}{\bibinfo{person}{Seyone Chithrananda},
  \bibinfo{person}{Gabriel Grand}, {and} \bibinfo{person}{Bharath Ramsundar}.}
  \bibinfo{year}{2020}\natexlab{}.
\newblock \showarticletitle{{ChemBERTa}: {Large}-{Scale} {Self}-{Supervised}
  {Pretraining} for {Molecular} {Property} {Prediction}}.
\newblock  (\bibinfo{year}{2020}).
\newblock
\urldef\tempurl%
\url{https://doi.org/10.48550/ARXIV.2010.09885}
\showDOI{\tempurl}
\newblock
\shownote{Publisher: arXiv}.


\bibitem[Cho et~al\mbox{.}(2017)]%
        {38}
\bibfield{author}{\bibinfo{person}{Hyejin Cho}, \bibinfo{person}{Wonjun Choi},
  {and} \bibinfo{person}{Hyunju Lee}.} \bibinfo{year}{2017}\natexlab{}.
\newblock \showarticletitle{A method for named entity normalization in
  biomedical articles: {Application} to diseases and plants}.
\newblock \bibinfo{journal}{\emph{BMC Bioinformatics}}  \bibinfo{volume}{18}
  (\bibinfo{year}{2017}).
\newblock
\urldef\tempurl%
\url{https://doi.org/10.1186/s12859-017-1857-8}
\showDOI{\tempurl}


\bibitem[Choi and Lee(2021)]%
        {3}
\bibfield{author}{\bibinfo{person}{Wonjun Choi} {and} \bibinfo{person}{Hyunju
  Lee}.} \bibinfo{year}{2021}\natexlab{}.
\newblock \showarticletitle{Identifying disease-gene associations using a
  convolutional neural network-based model by embedding a biological knowledge
  graph with entity descriptions}.
\newblock \bibinfo{journal}{\emph{PLOS ONE}} \bibinfo{volume}{16},
  \bibinfo{number}{10} (\bibinfo{year}{2021}), \bibinfo{pages}{1--27}.
\newblock
\urldef\tempurl%
\url{https://doi.org/10.1371/journal.pone.0258626}
\showDOI{\tempurl}
\newblock
\shownote{Publisher: Public Library of Science}.


\bibitem[Devlin et~al\mbox{.}(2018)]%
        {26}
\bibfield{author}{\bibinfo{person}{Jacob Devlin}, \bibinfo{person}{Ming-Wei
  Chang}, \bibinfo{person}{Kenton Lee}, {and} \bibinfo{person}{Kristina
  Toutanova}.} \bibinfo{year}{2018}\natexlab{}.
\newblock \showarticletitle{{BERT}: {Pre}-training of {Deep} {Bidirectional}
  {Transformers} for {Language} {Understanding}}.
\newblock  (\bibinfo{year}{2018}).
\newblock
\urldef\tempurl%
\url{https://doi.org/10.48550/ARXIV.1810.04805}
\showDOI{\tempurl}
\newblock
\shownote{Publisher: arXiv}.


\bibitem[Doğan et~al\mbox{.}(2014)]%
        {35}
\bibfield{author}{\bibinfo{person}{Rezarta~Islamaj Doğan},
  \bibinfo{person}{Robert Leaman}, {and} \bibinfo{person}{Zhiyong Lu}.}
  \bibinfo{year}{2014}\natexlab{}.
\newblock \showarticletitle{{NCBI} disease corpus: {A} resource for disease
  name recognition and concept normalization}.
\newblock \bibinfo{journal}{\emph{Journal of Biomedical Informatics}}
  \bibinfo{volume}{47} (\bibinfo{year}{2014}), \bibinfo{pages}{1--10}.
\newblock
\showISSN{1532-0464}
\urldef\tempurl%
\url{https://doi.org/10.1016/j.jbi.2013.12.006}
\showDOI{\tempurl}


\bibitem[Feng et~al\mbox{.}(2022)]%
        {18}
\bibfield{author}{\bibinfo{person}{Fan Feng}, \bibinfo{person}{Feitong Tang},
  \bibinfo{person}{Yijia Gao}, \bibinfo{person}{Dongyu Zhu},
  \bibinfo{person}{Tianjun Li}, \bibinfo{person}{Shuyuan Yang},
  \bibinfo{person}{Yuan Yao}, \bibinfo{person}{Yuanhao Huang}, {and}
  \bibinfo{person}{Jie Liu}.} \bibinfo{year}{2022}\natexlab{}.
\newblock \showarticletitle{{GenomicKB}: a knowledge graph for the human
  genome}.
\newblock \bibinfo{journal}{\emph{Nucleic Acids Research}}
  \bibinfo{volume}{51}, \bibinfo{number}{D1} (\bibinfo{year}{2022}),
  \bibinfo{pages}{D950--D956}.
\newblock
\showISSN{0305-1048}
\urldef\tempurl%
\url{https://doi.org/10.1093/nar/gkac957}
\showDOI{\tempurl}


\bibitem[Fukuda et~al\mbox{.}(1998)]%
        {22}
\bibfield{author}{\bibinfo{person}{Ken Fukuda}, \bibinfo{person}{Akihiro
  Tamura}, \bibinfo{person}{Tatsuhiko Tsunoda}, {and}
  \bibinfo{person}{Toshihisa Takagi}.} \bibinfo{year}{1998}\natexlab{}.
\newblock \showarticletitle{Toward information extraction: identifying protein
  names from biological papers}.
\newblock \bibinfo{journal}{\emph{Pacific Symposium on Biocomputing. Pacific
  Symposium on Biocomputing}} (\bibinfo{year}{1998}),
  \bibinfo{pages}{707--718}.
\newblock


\bibitem[Girju et~al\mbox{.}(2006)]%
        {41}
\bibfield{author}{\bibinfo{person}{Roxana Girju}, \bibinfo{person}{Adriana
  Badulescu}, {and} \bibinfo{person}{Dan Moldovan}.}
  \bibinfo{year}{2006}\natexlab{}.
\newblock \showarticletitle{Automatic {Discovery} of {Part}-{Whole}
  {Relations}}.
\newblock \bibinfo{journal}{\emph{Computational Linguistics}}
  \bibinfo{volume}{32}, \bibinfo{number}{1} (\bibinfo{year}{2006}),
  \bibinfo{pages}{83--135}.
\newblock
\showISSN{0891-2017}
\urldef\tempurl%
\url{https://doi.org/10.1162/coli.2006.32.1.83}
\showDOI{\tempurl}


\bibitem[Gábor et~al\mbox{.}(2018)]%
        {42}
\bibfield{author}{\bibinfo{person}{Kata Gábor}, \bibinfo{person}{Davide
  Buscaldi}, \bibinfo{person}{Anne-Kathrin Schumann}, \bibinfo{person}{Behrang
  QasemiZadeh}, \bibinfo{person}{Haïfa Zargayouna}, {and}
  \bibinfo{person}{Thierry Charnois}.} \bibinfo{year}{2018}\natexlab{}.
\newblock \showarticletitle{{SemEval}-2018 {Task} 7: {Semantic} {Relation}
  {Extraction} and {Classification} in {Scientific} {Papers}}.
  \bibinfo{publisher}{Association for Computational Linguistics},
  \bibinfo{address}{New Orleans, Louisiana}, \bibinfo{pages}{679--688}.
\newblock
\urldef\tempurl%
\url{https://doi.org/10.18653/v1/S18-1111}
\showDOI{\tempurl}


\bibitem[Kim et~al\mbox{.}(2019)]%
        {2}
\bibfield{author}{\bibinfo{person}{Jeongkyun Kim}, \bibinfo{person}{Jung-Jae
  Kim}, {and} \bibinfo{person}{Hyunju Lee}.} \bibinfo{year}{2019}\natexlab{}.
\newblock \showarticletitle{{DigChem}: {Identification} of
  disease-gene-chemical relationships from {Medline} abstracts}.
\newblock \bibinfo{journal}{\emph{PLOS Computational Biology}}
  \bibinfo{volume}{15}, \bibinfo{number}{5} (\bibinfo{year}{2019}),
  \bibinfo{pages}{1--16}.
\newblock
\urldef\tempurl%
\url{https://doi.org/10.1371/journal.pcbi.1007022}
\showDOI{\tempurl}
\newblock
\shownote{Publisher: Public Library of Science}.


\bibitem[Leaman et~al\mbox{.}(2015)]%
        {39}
\bibfield{author}{\bibinfo{person}{Robert Leaman}, \bibinfo{person}{Ritu
  Khare}, {and} \bibinfo{person}{Zhiyong Lu}.} \bibinfo{year}{2015}\natexlab{}.
\newblock \showarticletitle{Challenges in clinical natural language processing
  for automated disorder normalization}.
\newblock \bibinfo{journal}{\emph{Journal of Biomedical Informatics}}
  \bibinfo{volume}{57} (\bibinfo{year}{2015}), \bibinfo{pages}{28--37}.
\newblock
\showISSN{1532-0464}
\urldef\tempurl%
\url{https://doi.org/10.1016/j.jbi.2015.07.010}
\showDOI{\tempurl}


\bibitem[Lee et~al\mbox{.}(2019)]%
        {14}
\bibfield{author}{\bibinfo{person}{Jinhyuk Lee}, \bibinfo{person}{Wonjin Yoon},
  \bibinfo{person}{Sungdong Kim}, \bibinfo{person}{Donghyeon Kim},
  \bibinfo{person}{Sunkyu Kim}, \bibinfo{person}{Chan~Ho So}, {and}
  \bibinfo{person}{Jaewoo Kang}.} \bibinfo{year}{2019}\natexlab{}.
\newblock \showarticletitle{{BioBERT}: a pre-trained biomedical language
  representation model for biomedical text mining}.
\newblock \bibinfo{journal}{\emph{Bioinformatics}} \bibinfo{volume}{36},
  \bibinfo{number}{4} (\bibinfo{year}{2019}), \bibinfo{pages}{1234--1240}.
\newblock
\showISSN{1367-4803}
\urldef\tempurl%
\url{https://doi.org/10.1093/bioinformatics/btz682}
\showDOI{\tempurl}


\bibitem[Li et~al\mbox{.}(2018)]%
        {23}
\bibfield{author}{\bibinfo{person}{Jing Li}, \bibinfo{person}{Aixin Sun},
  \bibinfo{person}{Jianglei Han}, {and} \bibinfo{person}{Chenliang Li}.}
  \bibinfo{year}{2018}\natexlab{}.
\newblock \showarticletitle{A {Survey} on {Deep} {Learning} for {Named}
  {Entity} {Recognition}}.
\newblock  (\bibinfo{year}{2018}).
\newblock
\urldef\tempurl%
\url{https://doi.org/10.48550/ARXIV.1812.09449}
\showDOI{\tempurl}
\newblock
\shownote{Publisher: arXiv}.


\bibitem[Luo et~al\mbox{.}(2022)]%
        {8}
\bibfield{author}{\bibinfo{person}{Ling Luo}, \bibinfo{person}{Po-Ting Lai},
  \bibinfo{person}{Chih-Hsuan Wei}, \bibinfo{person}{Cecilia~N Arighi}, {and}
  \bibinfo{person}{Zhiyong Lu}.} \bibinfo{year}{2022}\natexlab{}.
\newblock \showarticletitle{{BioRED}: a rich biomedical relation extraction
  dataset}.
\newblock \bibinfo{journal}{\emph{Briefings in Bioinformatics}}
  \bibinfo{volume}{23}, \bibinfo{number}{5} (\bibinfo{year}{2022}).
\newblock
\showISSN{1477-4054}
\urldef\tempurl%
\url{https://doi.org/10.1093/bib/bbac282}
\showDOI{\tempurl}


\bibitem[Marcus et~al\mbox{.}(1993)]%
        {33}
\bibfield{author}{\bibinfo{person}{Mitchell~P Marcus},
  \bibinfo{person}{Beatrice Santorini}, {and} \bibinfo{person}{Mary~Ann
  Marcinkiewicz}.} \bibinfo{year}{1993}\natexlab{}.
\newblock \showarticletitle{Building a {Large} {Annotated} {Corpus} of
  {English}: {The} {Penn} {Treebank}}.
\newblock \bibinfo{journal}{\emph{Computational Linguistics}}
  \bibinfo{volume}{19}, \bibinfo{number}{2} (\bibinfo{year}{1993}),
  \bibinfo{pages}{313--330}.
\newblock
\urldef\tempurl%
\url{https://aclanthology.org/J93-2004}
\showURL{%
\tempurl}
\newblock
\shownote{Place: Cambridge, MA Publisher: MIT Press}.


\bibitem[Matentzoglu et~al\mbox{.}(2022)]%
        {20}
\bibfield{author}{\bibinfo{person}{Nicolas Matentzoglu},
  \bibinfo{person}{Damien Goutte-Gattat}, \bibinfo{person}{Shawn Zheng~Kai
  Tan}, \bibinfo{person}{James~P Balhoff}, \bibinfo{person}{Seth Carbon},
  \bibinfo{person}{Anita~R Caron}, \bibinfo{person}{William~D Duncan},
  \bibinfo{person}{Joe~E Flack}, \bibinfo{person}{Melissa Haendel},
  \bibinfo{person}{Nomi~L Harris}, \bibinfo{person}{William~R Hogan},
  \bibinfo{person}{Charles~Tapley Hoyt}, \bibinfo{person}{Rebecca~C Jackson},
  \bibinfo{person}{Hyeongsik Kim}, \bibinfo{person}{Huseyin Kir},
  \bibinfo{person}{Martin Larralde}, \bibinfo{person}{Julie~A McMurry},
  \bibinfo{person}{James~A Overton}, \bibinfo{person}{Bjoern Peters},
  \bibinfo{person}{Clare Pilgrim}, \bibinfo{person}{Ray Stefancsik},
  \bibinfo{person}{Sofia M~C Robb}, \bibinfo{person}{Sabrina Toro},
  \bibinfo{person}{Nicole~A Vasilevsky}, \bibinfo{person}{Ramona Walls},
  \bibinfo{person}{Christopher~J Mungall}, {and} \bibinfo{person}{David
  Osumi-Sutherland}.} \bibinfo{year}{2022}\natexlab{}.
\newblock \showarticletitle{Ontology {Development} {Kit}: a toolkit for
  building, maintaining and standardizing biomedical ontologies}.
\newblock \bibinfo{journal}{\emph{Database}}  \bibinfo{volume}{2022}
  (\bibinfo{year}{2022}).
\newblock
\showISSN{1758-0463}
\urldef\tempurl%
\url{https://doi.org/10.1093/database/baac087}
\showDOI{\tempurl}


\bibitem[Milošević and Thielemann(2023)]%
        {21}
\bibfield{author}{\bibinfo{person}{Nikola Milošević} {and}
  \bibinfo{person}{Wolfgang Thielemann}.} \bibinfo{year}{2023}\natexlab{}.
\newblock \showarticletitle{Comparison of biomedical relationship extraction
  methods and models for knowledge graph creation}.
\newblock \bibinfo{journal}{\emph{Journal of Web Semantics}}
  \bibinfo{volume}{75} (\bibinfo{date}{Jan.} \bibinfo{year}{2023}),
  \bibinfo{pages}{100756}.
\newblock
\urldef\tempurl%
\url{https://doi.org/10.1016/j.websem.2022.100756}
\showDOI{\tempurl}
\newblock
\shownote{Publisher: Elsevier BV}.


\bibitem[Neves et~al\mbox{.}(2010)]%
        {40}
\bibfield{author}{\bibinfo{person}{Mariana Neves},
  \bibinfo{person}{José-María Carazo}, {and} \bibinfo{person}{Alberto
  Pascual-Montano}.} \bibinfo{year}{2010}\natexlab{}.
\newblock \showarticletitle{Moara: {A} {Java} library for extracting and
  normalizing gene and protein mentions}.
\newblock \bibinfo{journal}{\emph{BMC bioinformatics}}  \bibinfo{volume}{11}
  (\bibinfo{year}{2010}), \bibinfo{pages}{157}.
\newblock
\urldef\tempurl%
\url{https://doi.org/10.1186/1471-2105-11-157}
\showDOI{\tempurl}


\bibitem[Noh and Kavuluru(2021)]%
        {9}
\bibfield{author}{\bibinfo{person}{Jiho Noh} {and} \bibinfo{person}{Ramakanth
  Kavuluru}.} \bibinfo{year}{2021}\natexlab{}.
\newblock \showarticletitle{Joint {Learning} for {Biomedical} {NER} and
  {Entity} {Normalization}: {Encoding} {Schemes}, {Counterfactual} {Examples},
  and {Zero}-{Shot} {Evaluation}}. In \bibinfo{booktitle}{\emph{{BCB} '21}}.
  \bibinfo{publisher}{Association for Computing Machinery},
  \bibinfo{address}{New York, NY, USA}.
\newblock
\urldef\tempurl%
\url{https://doi.org/10.1145/3459930.3469533}
\showDOI{\tempurl}
\newblock
\shownote{Journal Abbreviation: BCB '21}.


\bibitem[Pawar et~al\mbox{.}(2017)]%
        {24}
\bibfield{author}{\bibinfo{person}{Sachin Pawar}, \bibinfo{person}{Girish~K
  Palshikar}, {and} \bibinfo{person}{Pushpak Bhattacharyya}.}
  \bibinfo{year}{2017}\natexlab{}.
\newblock \showarticletitle{Relation {Extraction} : {A} {Survey}}.
\newblock  (\bibinfo{year}{2017}).
\newblock
\urldef\tempurl%
\url{https://doi.org/10.48550/ARXIV.1712.05191}
\showDOI{\tempurl}
\newblock
\shownote{Publisher: arXiv}.


\bibitem[Salzberg(2018)]%
        {1}
\bibfield{author}{\bibinfo{person}{Steven~L Salzberg}.}
  \bibinfo{year}{2018}\natexlab{}.
\newblock \showarticletitle{Open questions: {How} many genes do we have?}
\newblock \bibinfo{journal}{\emph{BMC Biology}} \bibinfo{volume}{16},
  \bibinfo{number}{1} (\bibinfo{date}{Aug.} \bibinfo{year}{2018}).
\newblock
\showISSN{1741-7007}
\urldef\tempurl%
\url{https://doi.org/10.1186/s12915-018-0564-x}
\showDOI{\tempurl}
\newblock
\shownote{Publisher: BioMed Central}.


\bibitem[Singh et~al\mbox{.}(2021)]%
        {4}
\bibfield{author}{\bibinfo{person}{Gurnoor Singh}, \bibinfo{person}{Evangelia~A
  Papoutsoglou}, \bibinfo{person}{Frederique Keijts-Lalleman},
  \bibinfo{person}{Bilyana Vencheva}, \bibinfo{person}{Mark Rice},
  \bibinfo{person}{Richard G~F Visser}, \bibinfo{person}{Christian W~B Bachem},
  {and} \bibinfo{person}{Richard Finkers}.} \bibinfo{year}{2021}\natexlab{}.
\newblock \showarticletitle{Extracting knowledge networks from plant scientific
  literature: potato tuber flesh color as an exemplary trait}.
\newblock \bibinfo{journal}{\emph{BMC Plant Biology}} \bibinfo{volume}{21},
  \bibinfo{number}{1} (\bibinfo{date}{April} \bibinfo{year}{2021}).
\newblock
\showISSN{1471-2229}
\urldef\tempurl%
\url{https://doi.org/10.1186/s12870-021-02943-5}
\showDOI{\tempurl}
\newblock
\shownote{Publisher: Springer Verlag}.


\bibitem[Smith et~al\mbox{.}(2008)]%
        {37}
\bibfield{author}{\bibinfo{person}{Larry~L Smith}, \bibinfo{person}{Lorraine~K
  Tanabe}, \bibinfo{person}{Rie Ando}, \bibinfo{person}{Cheng-Ju Kuo},
  \bibinfo{person}{I-Fang Chung}, \bibinfo{person}{Chun-Nan Hsu},
  \bibinfo{person}{Yu-Shi Lin}, \bibinfo{person}{Roman Klinger},
  \bibinfo{person}{C Friedrich}, \bibinfo{person}{Kuzman Ganchev},
  \bibinfo{person}{Manabu Torii}, \bibinfo{person}{Hongfang Liu},
  \bibinfo{person}{Barry Haddow}, \bibinfo{person}{Craig~A Struble},
  \bibinfo{person}{Richard~J Povinelli}, \bibinfo{person}{Andreas Vlachos},
  \bibinfo{person}{William~A Baumgartner}, \bibinfo{person}{Lawrence~E Hunter},
  \bibinfo{person}{Bob Carpenter}, \bibinfo{person}{Richard Tzong-Han Tsai},
  \bibinfo{person}{Hong-Jie Dai}, \bibinfo{person}{Feng Liu},
  \bibinfo{person}{Yifei Chen}, \bibinfo{person}{Chengjie Sun},
  \bibinfo{person}{Sophia Katrenko}, \bibinfo{person}{Pieter~W Adriaans},
  \bibinfo{person}{Christian Blaschke}, \bibinfo{person}{Rafael Torres},
  \bibinfo{person}{Mariana~L Neves}, \bibinfo{person}{Preslav Nakov},
  \bibinfo{person}{Anna Divoli}, \bibinfo{person}{Manuel Maña-López},
  \bibinfo{person}{Jacinto Mata}, {and} \bibinfo{person}{W~John Wilbur}.}
  \bibinfo{year}{2008}\natexlab{}.
\newblock \showarticletitle{Overview of {BioCreative} {II} gene mention
  recognition}.
\newblock \bibinfo{journal}{\emph{Genome Biology}}  \bibinfo{volume}{9}
  (\bibinfo{year}{2008}), \bibinfo{pages}{S2--S2}.
\newblock


\bibitem[Vaswani et~al\mbox{.}(2017)]%
        {12}
\bibfield{author}{\bibinfo{person}{Ashish Vaswani}, \bibinfo{person}{Noam
  Shazeer}, \bibinfo{person}{Niki Parmar}, \bibinfo{person}{Jakob Uszkoreit},
  \bibinfo{person}{Llion Jones}, \bibinfo{person}{Aidan~N Gomez},
  \bibinfo{person}{Lukasz Kaiser}, {and} \bibinfo{person}{Illia Polosukhin}.}
  \bibinfo{year}{2017}\natexlab{}.
\newblock \showarticletitle{Attention {Is} {All} {You} {Need}}.
\newblock  (\bibinfo{year}{2017}).
\newblock
\urldef\tempurl%
\url{https://doi.org/10.48550/ARXIV.1706.03762}
\showDOI{\tempurl}
\newblock
\shownote{Publisher: arXiv}.


\bibitem[Verma et~al\mbox{.}(2023)]%
        {29}
\bibfield{author}{\bibinfo{person}{Shilpa Verma}, \bibinfo{person}{Rajesh
  Bhatia}, \bibinfo{person}{Sandeep Harit}, {and} \bibinfo{person}{Sanjay
  Batish}.} \bibinfo{year}{2023}\natexlab{}.
\newblock \showarticletitle{Scholarly knowledge graphs through structuring
  scholarly communication: a review}.
\newblock \bibinfo{journal}{\emph{Complex \&amp; intelligent systems}}
  \bibinfo{volume}{9}, \bibinfo{number}{1} (\bibinfo{year}{2023}),
  \bibinfo{pages}{1059--1095}.
\newblock
\showISSN{2199-4536}
\urldef\tempurl%
\url{https://doi.org/10.1007/s40747-022-00806-6}
\showDOI{\tempurl}


\bibitem[Wang et~al\mbox{.}(2009)]%
        {10}
\bibfield{author}{\bibinfo{person}{Xinglong Wang}, \bibinfo{person}{Jun'ichi
  Tsujii}, {and} \bibinfo{person}{Sophia Ananiadou}.}
  \bibinfo{year}{2009}\natexlab{}.
\newblock \showarticletitle{Classifying {Relations} for {Biomedical} {Named}
  {Entity} {Disambiguation}}. \bibinfo{publisher}{Association for Computational
  Linguistics}, \bibinfo{address}{Singapore}, \bibinfo{pages}{1513--1522}.
\newblock
\urldef\tempurl%
\url{https://aclanthology.org/D09-1157}
\showURL{%
\tempurl}


\bibitem[Webster and Kit(1992)]%
        {32}
\bibfield{author}{\bibinfo{person}{Jonathan~J Webster} {and}
  \bibinfo{person}{Chunyu Kit}.} \bibinfo{year}{1992}\natexlab{}.
\newblock \showarticletitle{Tokenization as the {Initial} {Phase} in {NLP}}. In
  \bibinfo{booktitle}{\emph{{COLING} '92}}. \bibinfo{publisher}{Association for
  Computational Linguistics}, \bibinfo{address}{USA},
  \bibinfo{pages}{1106--1110}.
\newblock
\urldef\tempurl%
\url{https://doi.org/10.3115/992424.992434}
\showDOI{\tempurl}
\newblock
\shownote{Journal Abbreviation: COLING '92}.


\bibitem[Whetzel et~al\mbox{.}(2011)]%
        {19}
\bibfield{author}{\bibinfo{person}{Patricia Whetzel}, \bibinfo{person}{Natasha
  Noy}, \bibinfo{person}{Nigam Shah}, \bibinfo{person}{Paul Alexander},
  \bibinfo{person}{Csongor Nyulas}, \bibinfo{person}{Tania Tudorache}, {and}
  \bibinfo{person}{Mark Musen}.} \bibinfo{year}{2011}\natexlab{}.
\newblock \showarticletitle{{BioPortal}: {Enhanced} functionality via new {Web}
  services from the {National} {Center} for {Biomedical} {Ontology} to access
  and use ontologies in software applications}.
\newblock \bibinfo{journal}{\emph{Nucleic acids research}}
  \bibinfo{volume}{39} (\bibinfo{year}{2011}), \bibinfo{pages}{W541--5}.
\newblock
\urldef\tempurl%
\url{https://doi.org/10.1093/nar/gkr469}
\showDOI{\tempurl}


\bibitem[Wu et~al\mbox{.}(2019)]%
        {11}
\bibfield{author}{\bibinfo{person}{Ye Wu}, \bibinfo{person}{Ruibang Luo},
  \bibinfo{person}{Henry C~M Leung}, \bibinfo{person}{Hing-Fung Ting}, {and}
  \bibinfo{person}{Tak~Wah Lam}.} \bibinfo{year}{2019}\natexlab{}.
\newblock \showarticletitle{{RENET}: {A} {Deep} {Learning} {Approach} for
  {Extracting} {Gene}-{Disease} {Associations} from {Literature}}.
\newblock


\bibitem[Yang et~al\mbox{.}(2021a)]%
        {25}
\bibfield{author}{\bibinfo{person}{Jie Yang}, \bibinfo{person}{Soyeon~Caren
  Han}, {and} \bibinfo{person}{Josiah Poon}.} \bibinfo{year}{2021}\natexlab{a}.
\newblock \showarticletitle{A {Survey} on {Extraction} of {Causal} {Relations}
  from {Natural} {Language} {Text}}.
\newblock  (\bibinfo{year}{2021}).
\newblock
\urldef\tempurl%
\url{https://doi.org/10.48550/ARXIV.2101.06426}
\showDOI{\tempurl}
\newblock
\shownote{Publisher: arXiv}.


\bibitem[Yang et~al\mbox{.}(2021b)]%
        {31}
\bibfield{author}{\bibinfo{person}{Xi Yang}, \bibinfo{person}{Chengkun Wu},
  \bibinfo{person}{Goran Nenadic}, \bibinfo{person}{Wei Wang}, {and}
  \bibinfo{person}{Kai Lu}.} \bibinfo{year}{2021}\natexlab{b}.
\newblock \showarticletitle{Mining a stroke knowledge graph from literature}.
\newblock \bibinfo{journal}{\emph{BMC Bioinformatics}} \bibinfo{volume}{22},
  \bibinfo{number}{S10} (\bibinfo{date}{July} \bibinfo{year}{2021}).
\newblock
\showISSN{1471-2105}
\urldef\tempurl%
\url{https://doi.org/10.1186/s12859-021-04292-4}
\showDOI{\tempurl}
\newblock
\shownote{Publisher: Springer Nature}.


\bibitem[Zhu et~al\mbox{.}(2020)]%
        {15}
\bibfield{author}{\bibinfo{person}{Qian Zhu}, \bibinfo{person}{Dac-Trung
  Nguyen}, \bibinfo{person}{Ivan Grishagin}, \bibinfo{person}{Noel Southall},
  \bibinfo{person}{Eric Sid}, {and} \bibinfo{person}{Anne Pariser}.}
  \bibinfo{year}{2020}\natexlab{}.
\newblock \showarticletitle{An integrative knowledge graph for rare diseases,
  derived from the {Genetic} and {Rare} {Diseases} {Information} {Center}
  ({GARD})}.
\newblock \bibinfo{journal}{\emph{Journal of Biomedical Semantics}}
  \bibinfo{volume}{11} (\bibinfo{year}{2020}).
\newblock
\urldef\tempurl%
\url{https://doi.org/10.1186/s13326-020-00232-y}
\showDOI{\tempurl}


\bibitem[Zhu et~al\mbox{.}(2022)]%
        {30}
\bibfield{author}{\bibinfo{person}{Xian Zhu}, \bibinfo{person}{Yueming Gu},
  {and} \bibinfo{person}{Zhifeng Xiao}.} \bibinfo{year}{2022}\natexlab{}.
\newblock \showarticletitle{{HerbKG}: {Constructing} a {Herbal}-{Molecular}
  {Medicine} {Knowledge} {Graph} {Using} a {Two}-{Stage} {Framework} {Based} on
  {Deep} {Transfer} {Learning}}.
\newblock \bibinfo{journal}{\emph{Frontiers in Genetics}}  \bibinfo{volume}{13}
  (\bibinfo{year}{2022}).
\newblock
\showISSN{1664-8021}
\urldef\tempurl%
\url{https://doi.org/10.3389/fgene.2022.799349}
\showDOI{\tempurl}


\end{thebibliography}

\end{document}